\documentclass[conference,a4paper]{IEEEtran}
\IEEEoverridecommandlockouts

\usepackage[hidelinks]{hyperref}
\usepackage[cmex10]{amsmath}
\usepackage{amssymb,amsfonts}
\usepackage{dblfloatfix}

\usepackage[ruled,vlined]{algorithm2e}
\usepackage{graphicx}
\graphicspath{{Figures/PDF/}{Figures/PNG/}}

\usepackage{booktabs}
\usepackage{siunitx}
\usepackage[numbers,compress]{natbib}
\usepackage{texnames}
\usepackage{bm,bbm}
\usepackage{orcidlink}

\usepackage{subcaption}
\usepackage{cuted, tikz}
\usetikzlibrary{shapes, arrows.meta, positioning, calc, patterns, fit, backgrounds, shadows}

\begin{document}

\title{\uppercase{CRHT: a continuous regression hybrid transformer for vessel trajectory prediction with online cluster sampling}
}

\author{
	\IEEEauthorblockN{
        Alexander Schiøtz\orcidlink{0009-0002-1144-9885},
        Bertram Hage\orcidlink{0009-0005-5078-8363},
        Christian Rand\orcidlink{0000-0002-0574-0783},
        Felix Thomsen\orcidlink{0009-0009-8857-9267},
        Peder Heiselberg\orcidlink{0000-0002-8847-634X}
        \thanks{
        \copyright{} 2026 IEEE.  Personal use of this material is permitted.  Permission from IEEE must be obtained for all other uses, in any current or future media, including reprinting/republishing this material for advertising or promotional purposes, creating new collective works, for resale or redistribution to servers or lists, or reuse of any copyrighted component of this work in other works.
        }
        }
\IEEEauthorblockA{
\textit{Technical University of Denmark}\\
Anker Engelunds Vej 101
2800 Kongens Lyngby, Denmark\\
ph@space.dtu.dk
}
}

\maketitle
\begin{abstract}

Accurate vessel trajectory prediction is critical for maritime safety and anomaly detection, yet existing models often struggle with geographic bias and navigational realism. We propose the Continuous Regression Hybrid Transformer (CRHT), a deep learning framework designed to forecast vessel motion using Automatic Identification System (AIS) data. To mitigate spatial data imbalance, we introduce an online K-means cluster sampling strategy that ensures diverse exposure to rare maneuvers during training. Our hybrid architecture integrates 1D convolutional layers for local kinematic feature extraction with a multi-head attention mechanism for global temporal context. CRHT demonstrates superior performance in short-term forecasting, achieving the lowest errors at the 1-hour horizon. The results demonstrate that while discrete models provide high navigational stability over long horizons, CRHT offers an optimal balance of precision and maneuver tracking for real-time maritime surveillance.
\end{abstract}


\section{Introduction}
\label{sec:intro}

The Automatic Identification System (AIS) provides the spatio-temporal foundation for Maritime Domain Awareness (MDA). In critical waterways like the Danish Straits, accurate vessel trajectory prediction is vital for collision avoidance and anomaly detection. However, modeling AIS data is complicated by irregular sampling, sensor noise, and bias where models overfit to linear paths and fail during complex maneuvers.

Vessel trajectory prediction research has evolved from classical state-space models like the Kalman Filter \cite{Kalman1960}, which struggle with nonlinear maneuvers, to deep learning architectures. Recurrent Neural Networks (RNNs) and LSTMs \cite{zhang2022vessel} improved temporal modeling but are limited by the information bottleneck of fixed-size context vectors. Recently, Transformer-based architectures \cite{Nguyen2024, Wang2024} have set new benchmarks by using self-attention to capture long-range dependencies. While discrete generative models like the TrAISformer \cite{Nguyen2024} offer navigational realism through coordinate binning, they often suffer from quantization errors and high inference latency.

In this study we address the challenges of geographic data imbalance and regression precision by introducing the Continuous Regression Hybrid Transformer (CRHT). This work contributes a scalable MapReduce preprocessing pipeline, an online K-Means cluster sampling strategy to capture diverse maneuvers, and a hybrid CNN-Transformer architecture that integrates local kinematic extraction with global attention for robust trajectory forecasting.

\section{Data}
\label{sec:data}
The Danish Maritime Authority provides historic archives of vessel data from Denmark since 2006 \cite{DMA_AIS_Data}.
3 months of data in the period July 24th - October 24th 2025 was gathered. 
This dataset included 1,818,441,680 AIS messages from 35,349 unique vessels based on their unique Maritime Mobile Service Identity (MMSI) number. 

In this study we focused on larger vessels, namely tankers and cargo ships and their basic geographical and kinetic features of such as latitude, longitude, Speed over Ground (SOG) in knots, and Course over Ground (COG) in degrees. 
Additional information was available, such as vessel type and navigational status. 
However, these features lack data completeness, especially for smaller vessels.

The dataset was transformed from irregularly sampled and noisy single messages into regularly sampled coherent vessel trajectories.
First, we excluded erroneous messages outside our defined geographical boundary of $lat \in[54,59]$, $lon \in[5,17]$ and removed messages not conforming to $SOG \in [0,30]$ and $COG\in[0,360]$.
We then split the data by MMSI and perform further processing in parallel \cite{dean2008mapreduce}, reducing computation time and space while avoiding lossy simplification such as the otherwise popular Ramer–Douglas–Peucker (RDP) algorithm \cite{ramer1972iterative}.
Following \cite{Nguyen2024}, we defined a voyage as a sequence of messages with time gaps smaller than 2 hours, partitioned voyages longer than 20 hours, and discarded those shorter than 4 hours or containing fewer than 20 messages. 
Outliers where the calculated speed exceeded 40 knots were removed. 
Finally, we applied linear interpolation to resample the trajectories at 5-minute intervals. 
These steps reduced the dataset from 72,995 to 41,458 voyages spanning 1,203 unique vessels.

\section{Methodology}
\label{sec:method}
We frame vessel trajectory prediction task as a multivariate sequence-to-sequence regression problem. 
Given an observed sequence of historical states $X = \{x_1, \dots, x_T\}$ where $x_t\in\mathbb{R}^4$, we predict the future sequence $Y = \{y_{T+1}, \dots, y_{T+L}\}$ based on the feature set $F=\{\text{Latitude, Longitude, SOG, COG}\}$. 

We developed a Continuous Regression Hybrid Transformer (CRHT) for vessel trajectory prediction.
It performs continuous regression of coordinate displacements via three stages:

\noindent
\textbf{1. Local-Global Encoder:} The input sequence $X$ is processed by a 1D Convolutional block ($Conv1d \to ReLU \to Conv1d$) to extract local kinematic derivatives (velocity, acceleration) and smooth sensor noise. The result is augmented with sinusoidal positional encodings and passed to an $N$-layer Transformer Encoder, which uses multi-head self-attention to capture long-range dependencies across the 64-step history.

\noindent
\textbf{2. Query-Based Decoder:} Unlike recurrent architectures, we utilize a Transformer Decoder with $L$ learnable query embeddings $Q \in \mathbb{R}^{L \times d_{model}}$. These queries interact with the encoder's memory via multi-head \textit{cross-attention}, allowing the model to attend to specific historical maneuvers (e.g. turn entries) to predict future steps $t+1 \dots t+L$ in parallel.

\noindent
\textbf{3. Scaled Delta Learning:} To ensure stability, inputs are MinMax normalized to $[0,1]$, and the model regresses inter-step displacements $\Delta \hat{y}_t$. Predictions are denormalized during the forward pass and optimized using Huber Loss against scaled ground-truth targets to prevent gradient underflow:
\begin{equation}
\mathcal{L}_{TPTrans} = \text{Huber}(\Delta \hat{y}_{pred}, \Delta y_{true} \cdot \alpha)
\end{equation}
where $\alpha=100.0$ is a scaling factor applied to physical degree differences.\\

\begin{figure}[t]
    \centering
    \resizebox{\linewidth}{!}{
        \usetikzlibrary{shapes, arrows.meta, positioning, calc, patterns, fit, backgrounds, shadows}

\definecolor{encColor}{RGB}{225, 240, 255} 
\definecolor{decColor}{RGB}{255, 235, 225} 
\definecolor{dataColor}{RGB}{255, 255, 255} 
\definecolor{darkBlue}{RGB}{50, 80, 120}
\definecolor{darkRed}{RGB}{150, 50, 50}

\begin{tikzpicture}[
    >=Stealth,
    thick,
    font=\sffamily\footnotesize, 
    node distance=1.0cm,         
    vector/.style={
        rectangle, 
        draw=black!80, 
        fill=dataColor,
        minimum width=0.5cm, 
        minimum height=2.0cm, 
        pattern=grid, 
        pattern color=gray!30,
        inner sep=0pt,
        drop shadow={opacity=0.15}
    },
    embedding/.style={
        rectangle,
        draw=black!80,
        top color=gray!80, 
        bottom color=gray!50,
        minimum width=1.2cm,
        minimum height=0.4cm,
        rounded corners=1pt,
        drop shadow={opacity=0.15}
    },
    encModule/.style={
        rectangle, 
        draw=darkBlue, 
        top color=encColor!20, 
        bottom color=encColor, 
        rounded corners=3pt, 
        minimum width=2.0cm, 
        minimum height=1.0cm, 
        align=center,
        font=\sffamily\footnotesize\bfseries,
        drop shadow={opacity=0.2}
    },
    decModule/.style={
        rectangle, 
        draw=darkRed, 
        top color=decColor!20, 
        bottom color=decColor, 
        rounded corners=3pt, 
        minimum width=2.0cm, 
        minimum height=1.0cm, 
        align=center,
        font=\sffamily\footnotesize\bfseries,
        drop shadow={opacity=0.2}
    },
    op/.style={
        circle,
        draw=black!80,
        fill=white,
        inner sep=1pt,
        font=\scriptsize
    },
    connector/.style={
        ->,
        draw=black!70,
        rounded corners=3pt,
        thick
    }
]


\node[vector, label=below:Input $\mathbf{X}$] (input) at (0,0) {};

\node[encModule, right=1.2cm of input] (conv) {1D Conv\\Block};
\draw[connector] (input) -- (conv);

\node[embedding, right=0.8cm of conv, label=below:\scriptsize Feat. Emb.] (feat_emb) {};
\draw[connector] (conv) -- (feat_emb);

\node[encModule, right=1.2cm of feat_emb] (encoder) {Transf.\\Encoder};
\draw[connector] (feat_emb) -- node[midway, above, font=\scriptsize, inner sep=2pt, text=darkBlue] {+PE} (encoder);


\node[vector, above=3.0cm of feat_emb, label=below:Queries $\mathbf{Q}$] (queries) {};

\node[decModule] (decoder) at (encoder |- queries) {Transf.\\Decoder};

\draw[connector] (queries) -- node[midway, above, font=\scriptsize, inner sep=2pt, text=darkRed] {+PE} (decoder);

\draw[->, dashed, draw=black!60, thick] (encoder) -- node[midway, right, font=\scriptsize] {Memory} (decoder);


\node[decModule, right=1.0cm of decoder] (proj) {Linear\\Proj.};
\draw[connector] (decoder) -- (proj);

\node[vector, right=1.0cm of proj, label=below:Pred $\Delta \hat{\mathbf{y}}$] (pred) {};
\draw[connector] (proj) -- (pred);


\node[vector] (true) at (pred |- input) [label=below:True $\Delta \mathbf{y}$] {};

\node[op, right=0.5cm of true] (scale) {$\times \alpha$};
\draw[connector] (true) -- (scale);

\node[vector, right=0.5cm of scale, label=below:Scaled] (scaled) {};
\draw[connector] (scale) -- (scaled);

\coordinate (loss_pos) at ($(pred.east)!0.5!(scaled.east)$);

\node[circle, draw=red!80, dashed, fill=red!5, inner sep=2pt, align=center, right=1.0cm of loss_pos] (loss) {\scriptsize Huber\\\scriptsize Loss};

\draw[connector] (pred.east) -| (loss.north);
\draw[connector] (scaled.east) -| (loss.south);

\begin{scope}[on background layer]
    \node[fit=(input)(encoder)(feat_emb)(conv), rounded corners, fill=blue!5, draw=blue!10, inner sep=14pt, label={[anchor=north west, font=\footnotesize\bfseries, text=darkBlue, xshift=5pt, yshift=-1pt]north west:Encoder Stream}] (encBox) {};
    
    \node[fit=(queries)(decoder)(proj)(pred), rounded corners, fill=orange!5, draw=orange!10, inner sep=14pt, label={[anchor=north west, font=\footnotesize\bfseries, text=darkRed, xshift=5pt, yshift=-1pt]north west:Decoder Stream}] (decBox) {};
    
    \node[anchor=north east, font=\scriptsize\itshape, text=black!60] at (decBox.north east) {Future Steps $t+1 \dots t+L$};
\end{scope}

\end{tikzpicture} 
    }
    \caption{Architecture of the CRHT model. The dual-stream design separates the historical context encoding (bottom blue stream) from the future trajectory decoding (top orange stream). The decoder utilizes Positional Encoding (PE) and learnable queries combined with cross-attention to the encoder's memory to regress displacement deltas in parallel.}
    \label{fig:tptrans_arch}
\end{figure}
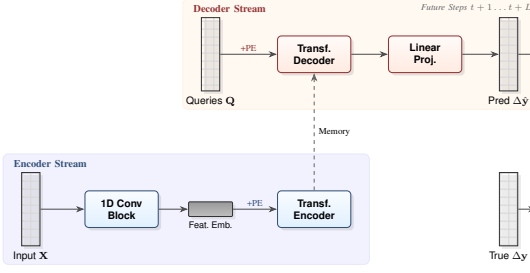

\subsection{Online K-Means Cluster Sampling}
Transformer-based sequence models incur a quadratic memory cost $\mathcal{O}(N^2)$ with respect to the input sequence length $N$, which renders direct training on multi-hour vessel trajectories infeasible with limited resources.
Consequently, long trajectories are truncated during training and fixed-length subsequences extracted to reduce memory cost.
However, uniform random sampling introduces a strong bias toward geographically dense and low-speed regions, as commercial vessels repeatedly traverse established shipping lanes and spend extended periods maneuvering near ports.
As a result, uniformly sampled subsequences overrepresent spatially redundant segments while underrepresenting rarer route deviations and sparsely trafficked areas.

\begin{figure}[t]
        \centering
        \includegraphics[width=\linewidth]{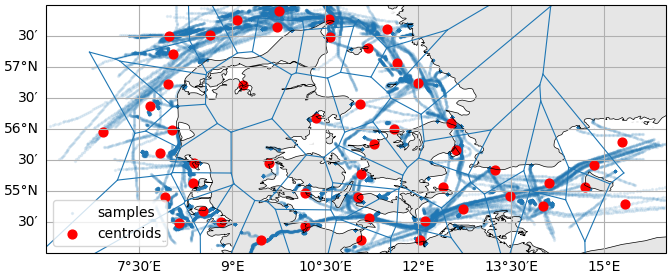}
        \caption{Voronoi tessellation \cite{du1999centroidal} of the coordinate space based on $k=50$ centroids (red dots). The blue lines represent the boundaries between clusters, ensuring that training samples are drawn from diverse geographical regions.}
        \label{fig:kmeans_voronoi}
\end{figure}

To mitigate this imbalance, we propose an online cluster-based sampling strategy based on spatial K-Means clustering \cite{macqueen1967multivariate}.
K-Means is a highly efficient clustering method with distributed implementations that makes it effective for huge AIS datasets \cite{balcan2013distributed}.
A K-Means algorithm with $k=50$ clusters is fitted on the latitude–longitude coordinates of AIS messages in the dataset.
Each AIS message is assigned a cluster index corresponding to its nearest centroid in Euclidean coordinate space (see Fig. \ref{fig:kmeans_voronoi}).

\begin{figure}[t]
    \centering
    \begin{subfigure}[a]{1.0\linewidth}
        \centering
        \includegraphics[width=\linewidth]{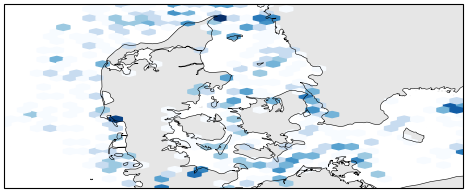}
        \caption{}
        \label{fig:kmeans_density1}
    \end{subfigure}

    \begin{subfigure}[b]{1.0\linewidth}
        \centering
        \includegraphics[width=\linewidth]{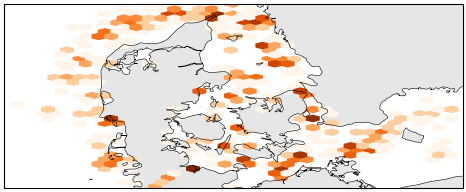}
        \caption{}
        \label{fig:kmeans_density2}
    \end{subfigure}

    \caption{Trajectory sampling start-point density. \textbf{Top:} Random sampling (blue). \textbf{Bottom:} K-Means sampling (orange).}
\end{figure}

Online (during training) sampling is performed by first selecting a trajectory, after which a cluster is drawn uniformly from the clusters overlapping that segment.
A starting index is then sampled uniformly over part of the sequence within the selected cluster, subject to the constraint that a contiguous subsequence can be formed.
This procedure enforces approximately uniform exposure across spatial clusters while preserving temporal continuity within each sampled subsequence (see Fig. \ref{fig:kmeans_density1} and \ref{fig:kmeans_density2}).

\section{Experiments}
As a non-learning baseline, we implemented a standard Kalman Filter assuming a Constant Velocity (CV) model based on the vessel's last known heading and speed \cite{Kalman1960,kalman}. 

We also implemented a CRHT using a Gated Recurrent Unit (GRU) decoder. However, it performed poorly compared to the Transformer decoder, which we attribute to the Information Bottleneck problem: the GRU is forced to compress the entire 64-step history into a single fixed-size context vector \cite{Bahdanau2015,Cho2014}. Consequently, the Transformer Decoder was selected for the final model. Furthermore, we evaluated different model capacities during training. 
We found that a medium-sized architecture offered the optimal trade-off between performance and generalization.

For the final CRHT implementation, we set the latent dimension $d_{model}=512$ with 8 attention heads, 8 encoder layers, and 8 decoder layers.

\subsection{Evaluation}
\label{sec:eval_strategy}
The dataset was partitioned into training, validation, and test sets in an 80:10:10 ratio split by MMSI to avoid data leakage and prevent the model from overfitting to individual vessels' movement patterns. We train using the AdamW optimizer \cite{loshchilov2017decoupled} with a batch size of 512 and an initial learning rate of $3 \times 10^{-4}$ on a single NVIDIA A100 (80GB) GPU \cite{dtu_hpc}. Training runs for a maximum of 200 epochs with early stopping (patience 50), utilizing a 5-epoch warm-up followed by validation-based learning rate decay (factor 0.5) triggered by loss stagnation.
During training we fix $T=64$ (past) and $L=12$ (future). 
All evaluated models benefitted from the online K-Means cluster sampling, providing a fair comparison of kinematic performance.

While \cite{Nguyen2024} employs a "Best-of-N" sampling strategy ($N=16$), selecting the best prediction post-hoc, we reject this as unsuitable for real-world maritime surveillance (e.g., Real-time trajectory prediction, port prediction, anomaly detection, and collision avoidance), where the true future is unknown at inference time. Instead, we enforce strict deterministic evaluation to assess reliability: \cite{Nguyen2024} uses greedy decoding ($T=0$) to select the most probable bin; CRHT regresses the deterministic mean of displacement deltas; and the Kalman Filter projects linearly without noise. 

To handle long horizons, we employ autoregressive generation: \cite{Nguyen2024} predicts step-by-step ($t \to t+1$), while CRHT operates block-autoregressively, forecasting full 1-hour horizons ($H=12$) iteratively by appending predictions to the history window.

We evaluate performance in kilometers (km) using the Haversine distance function $d_{hav}(\cdot)$. We report the Average Displacement Error (ADE), measuring the mean path deviation, and the Final Displacement Error (FDE), assessing destination accuracy at step $T+L$:
\begin{equation}
    ADE = \frac{1}{L} \sum_{t=1}^{L} d_{hav}(\hat{y}_{T+t}, y_{T+t})
\end{equation}
FDE is the geospatial distance between the predicted and ground truth positions at the final time step ($T+L$). This metric evaluates the model's destination forecasting accuracy.
     \begin{equation}
         FDE = d_{haversine}(\hat{y}_{T+L}, y_{T+L})
     \end{equation}

The mean trip length was $4.17$ hours.

\begin{table}[t]
	\centering
	\caption{Mean FDE and ADE prediction errors in kilometers (km). \textbf{Bold} indicates the best performance in each category.}
	\label{tab:traj-metrics}
	\small
	\setlength{\tabcolsep}{4pt}
	\begin{tabular}{l 
		S[table-format=1.2] S[table-format=2.2]
		S[table-format=1.2] S[table-format=2.2]}
		\toprule
		& \multicolumn{2}{c}{\textbf{1 Hour Horizon}} 
		& \multicolumn{2}{c}{\textbf{2 Hour Horizon}} \\
		\cmidrule(lr){2-3} \cmidrule(lr){4-5}
		\textbf{Model} 
		& {\textbf{ADE}} & {\textbf{FDE}} 
		& {\textbf{ADE}} & {\textbf{FDE}} \\
		\midrule
		Kalman Filter 
		& 1.58 & 3.60
		& 3.70 & 8.86 \\
		TrAISformer 
		& 1.38 & 2.41
		& 2.62 & \textbf{ 5.42}  \\
		CRHT 
		& \textbf{1.21} & \textbf{ 2.07}
		& \textbf{2.55} & 5.85 \\
		\bottomrule
	\end{tabular}
\end{table}

\section{Results \& Discussion}
\label{sec:results}

Table \ref{tab:traj-metrics} reports trajectory prediction ADE and FDE across 1-hour and 2-hour horizons. The Continuous Regression Hybrid Transformer (CRHT) consistently achieves the strongest overall performance profile across both horizons.

At a 1-hour prediction horizon, CRHT achieves the lowest mean ADE (1.21 km) and the lowest mean FDE (2.07 km) among all evaluated models. 
At the 2-hour horizon, CRHT maintains the lowest mean ADE (2.55 km). For FDE, CRHT achieves a mean error of 5.85 km, which is slightly higher than TrAISformer (5.42 km).
Across horizons, CRHT demonstrates a consistent reduction in displacement errors relative to classical filtering and discrete generative baselines.

\begin{figure}[t]
        \centering
        \includegraphics[width=\linewidth]{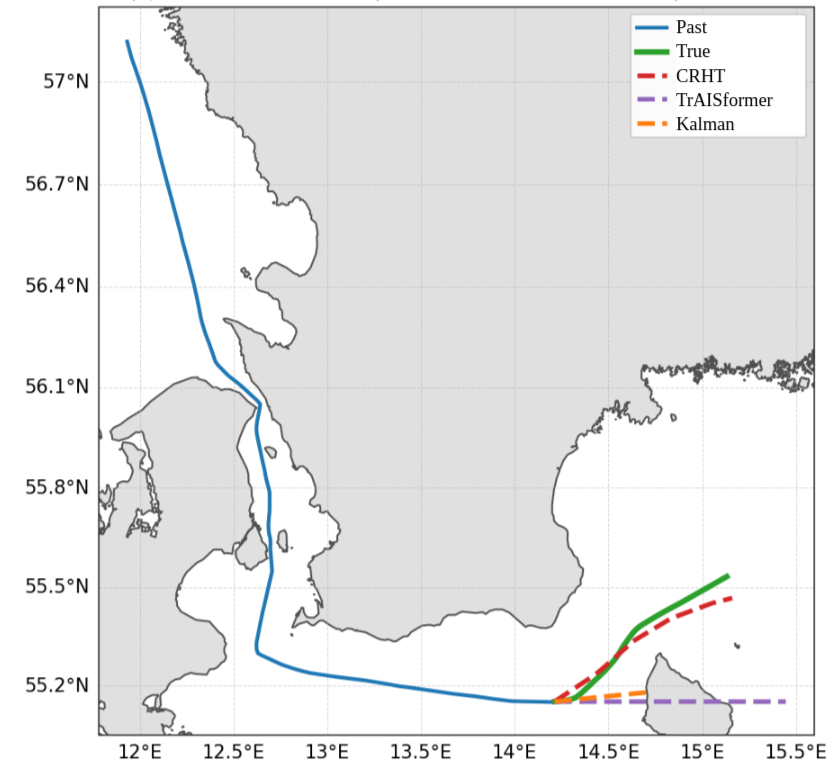}
        \caption{Vessel trajectory turning north around Bornholm.}
        \label{fig:traj_good}
\end{figure}

\begin{figure}[t]
        \centering
        \includegraphics[width=\linewidth]{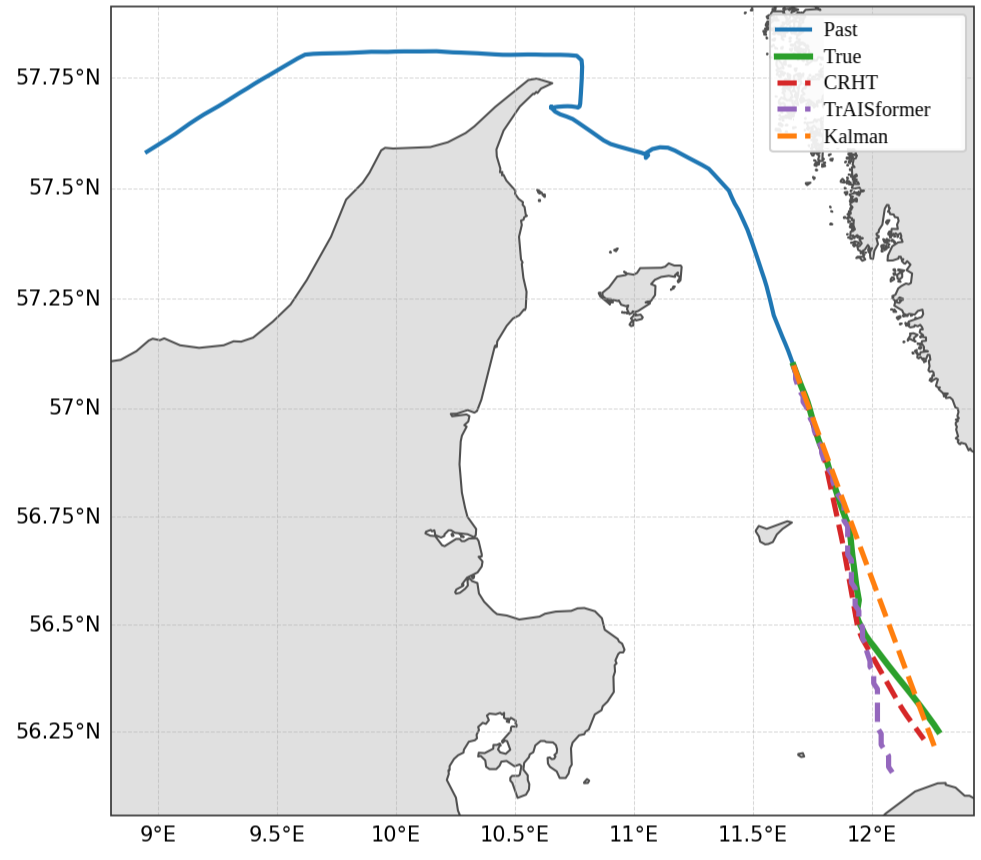}
        \caption{Vessel trajectory with a slight curve.}
        \label{fig:traj_ok}
\end{figure}

\begin{figure}[t]
        \centering
        \includegraphics[width=.5\linewidth]{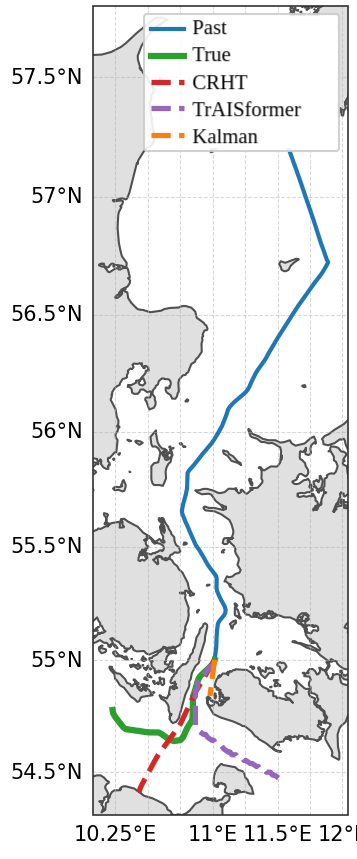}
        \caption{Vessel trajectory in an ambiguous scenario.}
        \label{fig:traj_bad}
\end{figure}

Fig. \ref{fig:traj_good} showcases a difficult prediction scenario in which a vessel, following a long and nearly linear trajectory south of Sweden, executes a pronounced northward turn around Bornholm. In this setting, CRHT predicts a trajectory that follows the northward maneuver and remains feasible over the full forecast horizon, closely matching the true vessel path. By contrast, both the Kalman Filter and TrAISformer extrapolate the preceding motion pattern and project the trajectory into the island of Bornholm, reflecting an inability to represent the observed change in heading in this scenario.

In Fig. \ref{fig:traj_ok}, CRHT produces a trajectory that largely aligns with the ground truth, with only a small spatial deviation. The Kalman Filter continues along a straight-line path and achieves the lowest FDE in this instance, but its trajectory does not capture the executed maneuver and deviates from the actual vessel path. This example illustrates that low point-wise error metrics may coincide with qualitatively inconsistent motion representations.

Fig. \ref{fig:traj_bad} depicts a highly ambiguous scenario in which none of the evaluated models accurately predict the realized vessel trajectory. CRHT and the Kalman Filter diverge early from the ground truth, while TrAISformer produces a trajectory that would be realistic if the vessel had taken an eastward turn instead of the executed westward maneuver, effectively "hallucinating" an alternative path. This example illustrates the challenge of modeling trajectories under high uncertainty, where multiple plausible routes exist and a model may generate a navigationally valid path that does not correspond to the observed behavior.

\section{Conclusion}
\label{sec:conclusion}
We introduced and evaluated the Continuous Regression Hybrid Transformer (CRHT) for AIS-based vessel trajectory prediction. CRHT achieves the lowest mean displacement errors across short- and intermediate-term horizons while producing smooth, continuous trajectories that handle nonlinear maneuvers. The results demonstrate its ability to follow complex paths realistically, in contrast to classical filtering and discrete generative models that either fail on sharp turns or generate plausible but unexecuted trajectories. By combining predictive accuracy with navigational feasibility, CRHT provides a robust framework for maritime trajectory forecasting under both typical and uncertain conditions.

All code for this study is available \href{https://github.com/alexander9908/AIS-MDA}{here}

\small
\bibliographystyle{IEEEtranN}
\bibliography{references}

\end{document}